\documentclass[conference]{IEEEtran}
\IEEEoverridecommandlockouts
\usepackage{cite}
\usepackage{amsmath,amssymb,amsfonts}
\usepackage{algorithmic}
\usepackage{graphicx,hhline,ragged2e}
\usepackage{textcomp,hyperref,adjustbox}
\usepackage{xcolor}
\usepackage{array}
\newcommand\ChangeRT[1]{\noalign{\hrule height #1}}
\def\BibTeX{{\rm B\kern-.05em{\sc i\kern-.025em b}\kern-.08em
    T\kern-.1667em\lower.7ex\hbox{E}\kern-.125emX}}
\begin{document}

\title{Curious Meta-Controller: Adaptive Alternation between Model-Based and Model-Free Control 
in Deep Reinforcement Learning
\\
}

\author{\IEEEauthorblockN{Muhammad Burhan Hafez, Cornelius Weber, Matthias Kerzel and Stefan Wermter }
\IEEEauthorblockA{\textit{Knowledge Technology, Department of Informatics, University of Hamburg, Germany}\\
\{hafez, weber, kerzel, wermter\}@informatik.uni-hamburg.de}
}

\maketitle

\begin{abstract}
Recent success in deep reinforcement learning for continuous control has been dominated by model-free approaches which, unlike model-based approaches, do not suffer from representational limitations in making assumptions about the world dynamics and model errors inevitable in complex domains. However, they require a lot of experiences compared to model-based approaches that are typically more sample-efficient. We propose to combine the benefits of the two approaches by presenting an integrated approach called Curious Meta-Controller. Our approach alternates adaptively between model-based and model-free control using a curiosity feedback based on the learning progress of a neural model of the dynamics in a learned latent space. We demonstrate that our approach can significantly improve the sample efficiency and achieve near-optimal performance on learning robotic reaching and grasping tasks from raw-pixel input in both dense and sparse reward settings.
\end{abstract}


\section{Introduction}
Deep Reinforcement Learning (RL) enables artificial agents to learn through trial and error a direct mapping from a raw sensory input to a raw motor output that results in an optimal control behavior for achieving a desired task. It has recently shown a great success across different domains, exceeding human performance in playing Atari games \cite{Mni15} and allowing the acquisition of complex robotic manipulation skills \cite{Lev16}. \par

One major issue, however, with the current deep RL algorithms is their poor sample efficiency, which becomes particularly problematic in robotic control where real-time constraints and noisy observations are common. Moreover, it is desirable for the agent to learn from sparse rewards that eliminate the need for complex and biased reward shaping. This poses a great challenge, because the agent lacks the important feedback on how to adjust its behavior before receiving a reward, further reducing the sample efficiency of the learning algorithm.\par

To address this issue, some approaches focus on how to make efficient use of the experience samples stored in a replay memory. For example, Schaul et al. \cite{Sch16} propose to sample experiences according to a priority based on their temporal-difference error instead of using uniform random sampling. In a more recent work, Andrychowicz et al. \cite{And17} show that replaying an episode with a different goal than the one given to the agent in a multi-goal, sparse reward setting greatly improves the sample efficiency, even more than when using a shaped reward function. However, their approach is based on a strong assumption that the goal is always separable from the state in order to use goal-conditioned reward and value functions, which is not applicable to domains where the goal representation is embedded in a raw-pixel observation and thus cannot be given as a separate input channel.\par


\begin{figure}[t]
	\centering
		\includegraphics[width=3in,height=2.5in]{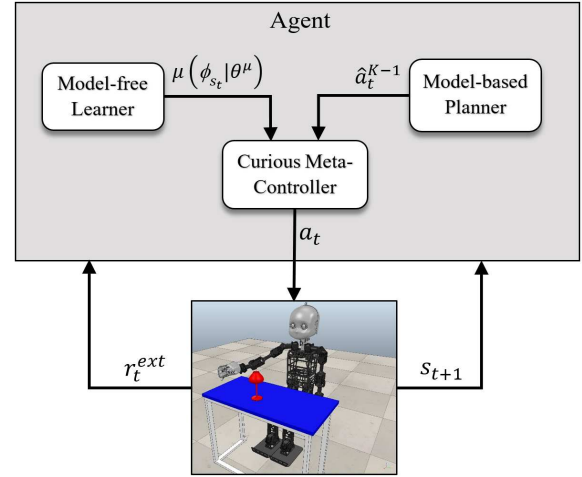}
		\caption{Curious Meta-Controller (CMC): At each timestep \textit{t}, CMC uses the adaptive learning progress $LP_{t-1}$  to decide which controller to query for an action. If $LP_{t-1}$ is positive, CMC queries the model-based planner which then performs planning in the learned latent space. After \textit{K} optimization iterations performed on an initial action plan from the model-free learner, the optimal plan's first action $\hat{a}_{t}^{K-1}$  is sent to the environment with exploration noise. If $LP_{t-1}$ is negative, CMC queries the actor neural network of the model-free learner for its estimate of the optimal action $ \mu(\phi_{s_{t}} \vert \theta^{\mu}) $ which is then sent to the environment with exploration noise.}
		\label{fig:1}
\end{figure}


\par

  Other approaches focus on the exploration problem itself to provide more efficient alternatives for collecting experience samples than the commonly used random exploration. Scaling count-based exploration strategies, which are limited to tabular representations of the environment, to large RL tasks with continuous, high-dimensional environments is one attempt in this direction \cite{Ost17},\cite{Tan17}. While these works use state-visitation counts to generate an exploration bonus for visiting novel states, works on intrinsic motivation and artificial curiosity use a self-generated reward signal based on the predictability of future states by a forward dynamics model to direct the exploration from highly to less predictable regions of the state space, especially in the absence of any extrinsic rewards. Examples of intrinsic reward functions include model prediction error \cite{Stadie15}, \cite{Pat17}, model learning progress \cite{Sch10,Got13,Haf18,Haf181}, change in policy value \cite{Haf15}, and information-theoretic-based dynamics uncertainty \cite{Hou16}. Although intrinsic motivation approaches provide an active exploration that enhances the sample efficiency of RL in sparse reward settings, the useful information the learned dynamics model offers is almost only employed to compute the intrinsic reward without exploiting it in the model-based learning of value and policy functions.\par
Using predictive models to accelerate RL is very appealing, primarily because they help minimize the expensive interactions with the real world by allowing to hallucinate experiences and do offline planning. However, in complex domains, they suffer from inevitable approximation errors that quickly compound when planning with the model and lead to inaccurate and useless long-term predictions. This has made model-based methods unable to match the success of model-free methods in deep RL for large-scale problems. In an effort to reduce the effect of the compounding prediction error of the learned model during planning, Talvitie \cite{Tal17} proposes a model-based RL algorithm where the model is trained via hallucinated replay to predict the next world state, given its own predictions as input, continually correcting itself. The algorithm has a theoretical guarantee on the error bound of the value of the target policy, but is limited to deterministic environments.\par

In this paper, we introduce the \textit{Curious Meta-Controller} (CMC), a novel intrinsically motivated meta-control approach for exploration that adaptively alternates between model-based planning and model-free RL (Fig. \ref{fig:1}). The alternation is controlled online via a curiosity signal based on the learning progress of an evolving dynamics model. In contrast to other related works, our approach takes the reliability of the learned model into account before using it for planning. In our approach, both the model-based planner and the model-free learner are mutually improving, since the model-free learner can give the planner a good initial action sequence and the model-based planner can give the learner a more informed exploratory action. CMC can be combined with any off-policy RL algorithm with minimal changes and is in line with findings from neuroscience on the dual-system approach to human decision-making. We evaluate popular deep continuous-action RL algorithms with and without CMC and show that CMC improves the sample efficiency and achieves better performance. \par

\section{Related Work}
\textbf{Neural models of hybrid control: }Interest has been growing recently to combine the advantages of model-free and model-based approaches, inspired by Dyna-Q, one of the earliest works of this kind, which is based on Q-learning but trained on both real and model-generated experiences \cite{Sut90}. For example, Nagabandi et al. \cite{Nag18} suggest that a trained model-based controller can be used to initialize the action policy of a model-free learner to help the latter be more sample-efficient. To address model imperfection, real samples resulting from executing the policy of the model-based controller are used in combination with those from random trajectories on which the dynamics model was trained to refit the model, reducing the distribution mismatch between random and controller-generated transitions. The use of a model predictive controller based on random-sampling however limits the applicability of the approach to low-dimensional action spaces and short planning horizons. Racanière et al. \cite{Web17}, on the other hand, handle the inaccurate predictions of an environment model by encoding rollouts of the imagined observations from the model with a recurrent neural network. The encoded rollouts are then concatenated and fed as an additional input to the model-free agent. Unlike other approaches, Kalweit and Boedecker \cite{Kal17} augment the model-free agent with model-generated imaginary samples only when there is a high uncertainty in the agent's predictions of its state-action values. While the approach is empirically shown to improve the efficiency of learning continuous control policies, it does not take into account prediction errors of the model. Gu et al. \cite{GuS16} also use imagination rollouts generated by a learned model to augment the buffer of real transitions and speed up model-free learning. They iteratively refit a linear model to a number of recent real rollouts. The model then generates short imagination rollouts from states sampled from the rollouts on which the model was trained. While this is an efficient method to learn a world model and involves less model bias, the model learned is not expressive enough to generate good rollouts in control tasks from raw-pixel input.\par

In a very different study, Srinivas et al. \cite{Sri18} find that learning a state representation and a dynamics model that improve gradient-descent planning based on a set of training demonstrations rather than optimize auxiliary objectives leads to more successful action plans. They show that the distance to a target image encoded with the learned representation can be effectively used as a reward for a model-free RL agent in visuomotor control tasks. The approach however requires expert demonstrations to be available for training. \par

More recently, a control architecture was proposed that includes an arbitrator used to switch between habitual and planning systems by choosing between an action predicted by an actor of an actor-critic model and that predicted by an inverse dynamics model \cite{Far18}. The arbitration is managed by the reward prediction error and favors the actor's prediction if the error at the previous timestep is below a predefined threshold. The approach does not consider imperfect model predictions and is applied to a significantly low-dimensional state space.\par

As opposed to explicitly learning a dynamics model, Pong et al. \cite{Pon18} propose a type of goal-conditioned value function called Temporal Difference Model (TDM) that implicitly learns a dynamics model and uses it for optimal control. In their approach, transitions collected off-policy are sampled from a replay buffer and relabeled with new, randomly sampled goal states and time horizons which the TDM uses as input along with the state-action pair. The TDM is learned model-free and updated to be the negative distance between the newly visited and goal states if the horizon is zero or, otherwise, to be the approximate TDM value after decrementing the horizon and advancing the state. The information the TDM provides on the closeness to the goal after a given number of actions makes it resemble a model. Despite achieving high sample efficiency by relabeling collected transitions with several goals and horizons, the approach has not been applied to learning from raw-pixel input but only to learning from simple low-dimensional observations.\par

\textbf{Dual-system decision-making in neuroscience: }Neuroscience studies on choice behavior have presented different hypotheses on how habitual (model-free) and planning (model-based) systems control human sequential decision-making. A study by Daw \cite{Daw15} argues for a deliberative planning system in which a learned model of the task is used to exhaustively search the decision tree until the goal is reached, while the habits are formed based on the expected long-term reward of an action, obtained on completion of the tree search. In contrast, Cushman and Morris \cite{Cus15} argue for a different hybrid control model where (sub-) goals are first chosen with model-free learning and then aimed at with model-based planning.\par

Keramati et al. \cite{Ker16} show behavioral evidence suggesting that the brain integrates habits in terms of learned estimates of future consequences of the current actions into depth-limited planning, proposing an integrative plan-until-habit framework. In the framework, the world is simulated up to a certain depth, which decreases with increased time pressure, and then the habitual values are exploited, as opposed to either following pure habitual or pure planning strategies. \par

In contrast, Kool et al. \cite{Koo18} propose that the arbitration between model-free and model-based control is driven by a cost-benefit trade-off and not by the cognitive ability to plan. They hypothesize that the brain estimates the expected value of using each of the two control systems during choice but then decreases that of the model-based proportional to its cognitive cost. This was supported by an observation that participants with even an accurate internal model of a decision-making task and an extended response time used less model-based control when its estimated reward advantage was low.\par

While these studies provide strong evidence for the dual-system approach to decision-making that is distinguishable neurally and behaviorally and can be utilized in more realistic computational models, they almost always assume a perfect internal model of the task. To relax this assumption, an intrinsic measure of the reliability of predictions of a learned model needs to be incorporated into the behavioral control system. This is most likely to guide the behavior to improve the learned model and eventually lead to a better hybrid control system.\par

\section{Background}
\subsection{Reinforcement Learning}
We consider a standard RL problem where an agent interacts with a fully observable environment using a policy to maximize accumulated future reward. An environment consists of a state space \textit{S}, an action Space \textit{A}, a reward function  \( r : S  \times A \rightarrow \mathbb{R} \), a dynamics model  \( p \left( s_{t+1}  \vert  s_{t}, a_{t} \right)  \), and a discount factor  \(  \gamma   \in   \left[ 0,1 \right]  \). A policy  \(  \pi  : S \rightarrow P \left( A \right)  \)  is a mapping from states to probability distribution over actions.\par

At each timestep \textit{t}, the agent takes an action  \( a_{t}  \sim   \pi   \left( s_{t} \right)  \)  and receives a reward  \( r_{t}= r  \left( s_{t}, a_{t} \right)  \)  while the environment transitions into a new state  \( s_{t+1}  \sim  p \left(  \cdot   \vert s_{t}, a_{t} \right)  \). A discounted sum of future rewards defines the return  \( R_{t}=  \sum _{i=t}^{T-1} \gamma ^{i-t} r \left( s_{i},a_{i} \right)  \). The goal is to maximize the expected return  \( \mathbb{E}_{s_{0} \sim S_{0}} \left[ R_{0}  \vert  s_{0} \right]  \), where  \( S_{0} \subseteq S \)  is a set of initial states. \par

An action-value (or \(Q-\)) function is defined as  \( Q^{ \pi } \left( s_{t}, a_{t} \right) = \mathbb{E}  \left[ R_{t}  \vert  s_{t}, a_{t} \right]  \), and the optimal policy  \(  \pi ^{\ast} \)  then satisfies  \( Q^{ \pi ^{\ast}} \left( s,a \right)   \geq  Q^{ \pi } \left( s,a \right),  \forall  \left( s,a \right)  \in S  \times A \). In deep RL and when the model is not available, the optimal \textit{Q}-function is approximated by a neural network with parameters  \(  \theta ^{Q} \)  and trained to minimize the loss  \( \mathcal{L} \)  between the target value  \( y_{t}= r \left( s_{t},a_{t} \right) +  \gamma max_{a}Q \left( s_{t+1},a  \vert   \theta ^{Q} \right)  \)  and the current \textit{Q}-estimate:\par

\begin{equation}
\mathcal{L}  \left(  \theta ^{Q} \right) =  \left( y_{t} - Q \left( s_{t},a_{t}  \vert   \theta ^{Q} \right)  \right) ^{2}
\end{equation}\par

In RL, actor-critic methods are well suited for continuous action spaces, since they learn a policy and a value function simultaneously. Of particular interest are the off-policy actor-critic methods since they allow for integrating exploratory actions from another controller, such as \textit{Deep Deterministic Policy Gradient} (DDPG) \cite{Lil15} and \textit{Continuous Actor-Critic Learning Automaton} (CACLA) \cite{Van12}.\par

\subsection{DDPG}

DDPG is a model-free RL algorithm that learns a deterministic target policy  \(  \mu  : S \rightarrow A \)  while acting according to a stochastic behavior policy (e.g. random exploration noise added to  \(  \mu  \)). DDPG approximates the policy function  \(  \mu  \)  and the \textit{Q}-function using two neural networks: an actor  \(  \mu  \left(  \cdot  \vert   \theta ^{ \mu } \right)  \)  and a critic  \( Q \left(  \cdot, \cdot  \vert  \theta ^{Q} \right)  \)  with parameters  \(  \theta ^{ \mu } \)  and  \(  \theta ^{Q} \)  respectively. The target values for the training use slowly updated actor and critic target networks  \(  \mu ^{'} \) and \(Q^{'}\), parameterized by  \(  \theta ^{ \mu'} \) and  \(  \theta ^{Q'} \) respectively. This stabilizes the learning, as previously found in \cite{Mni15}.  At each update step, a minibatch of \textit{n} experiences is randomly sampled from an experience replay memory. The critic is updated to minimize the loss  \( \mathcal{L}  \left(  \theta ^{Q} \right) = \frac{1}{n}  \sum _{i}^{} \left( y_{i}- Q \left( s_{i}, a_{i}  \vert   \theta ^{Q} \right)  \right) ^{2} \), where  \(  y_{i}= r_{i} +  \gamma Q' \left( s_{i+1},~  \mu ' \left( s_{i+1}  \vert   \theta ^{ \mu '} \right)  \vert   \theta ^{Q'} \right)  \). The actor is then updated by minibatch gradient ascent on the \textit{Q}-function with respect to  \(  \theta ^{ \mu } \), following the policy gradient:
\begin{align}
\begin{split}
&\triangledown _{ \theta ^{ \mu }}\frac{1}{n}  \sum _{i}^{}Q \left( s_{i},  \mu  \left( s_{i} \right)  \vert   \theta ^{Q} \right)  \\
&= \frac{1}{n}  \sum _{i}^{}\triangledown _{a}Q \left( s,a  \vert   \theta ^{Q} \right)  \vert _{s= s_{i}, a=  \mu  \left( s_{i} \right) } \triangledown _{ \theta ^{ \mu }} \mu  \left( s  \vert   \theta ^{ \mu } \right)  \vert _{s= s_{i}}
\end{split}
\end{align}

The target network parameters  \(  \theta ^{ \mu^{'}} \)  and  \(  \theta ^{Q^{'}} \)  are updated slowly towards their corresponding network parameters  \(  \theta ^{ \mu } \)  and  \(  \theta ^{Q} \)  :\par

 \[  \theta ^{Q^{'}} \leftarrow   \tau  \theta ^{Q}+  \left( 1- \tau \right)  \theta ^{Q^{'}} \]  \[  \theta ^{ \mu ^{'}} \leftarrow   \tau  \theta ^{ \mu }+  \left( 1- \tau \right)  \theta ^{ \mu ^{'}}, \text{ with }  \tau  \ll 1. \] 
\par

\subsection{CACLA}
Similar to DDPG, CACLA is a model-free actor-critic algorithm. However, the policy here is updated only towards an exploratory action that improves the value estimate of CACLA's current estimate of the best action. CACLA can operate on-policy by approximating a state-value function and learning a stochastic target policy or off-policy by approximating a \textit{Q}-function and learning a deterministic target policy. In the latter case, the critic  \( Q \left(  \cdot, \cdot  \vert  \theta ^{Q} \right)  \)  is updated as in DDPG, while the actor  \(  \mu  \left(  \cdot  \vert   \theta ^{ \mu } \right)  \)  is updated towards the action  \( a_{t} \)  (generated by an arbitrary behavior policy) of experience samples for which the advantage estimator \( \hat{A}_{t} \)  is positive by gradient descent on the loss  \( \frac{1}{2}  \left( a_{t}-  \mu  \left( s_{t} \vert  \theta ^{ \mu } \right)  \right) ^{2} \) :
 \[ \text{If } \hat{A}_{t}>0 : \theta ^{ \mu }  \leftarrow   \theta ^{ \mu }+  \alpha   \left( a_{t}-  \mu  \left( s_{t} \vert  \theta ^{ \mu } \right)  \right) \triangledown _{ \theta ^{ \mu }} \mu  \left( s_{t} \vert  \theta ^{ \mu } \right)  \]
where  \(  \alpha  \in  \left[ 0,1 \right]  \)  is the learning rate and  \( \hat{A}_{t} \) is the observed advantage of action \(a_{t}\), which is the difference between the value estimate of the current best action and the observed value of the action \( a_{t}\) :  \( \hat{A}_{t}= r_{t} +  \gamma Q \left( s_{t+1},~  \mu  \left( s_{t+1}  \vert   \theta ^{ \mu } \right)  \vert   \theta ^{Q} \right) - Q \left( s_{t},  \mu  \left( s_{t} \vert  \theta ^{ \mu } \right)   \vert   \theta ^{Q} \right)  \). The reason for the conditional update is that when an exploratory action is found to have a value greater than the critic's estimate of the value of the best action (\( \hat{A}_{t}   > 0\)), it is most likely to yield higher future rewards and so the policy is updated towards that action. \par

\section{Curious Meta-Controller}

In this section, we present our Curious Meta-Controller (CMC) for adaptive alternation between model-based and model-free control. CMC consists of two interacting and mutually improving components: the model-based planner and the model-free learner.\par

\subsection{Model-based planner}
\label{sec4a}
In our approach, we train a neural network dynamics model that takes as input the state encoding and the action for the current timestep and predicts the state encoding and the environment reward for the next timestep. The model parameters are periodically updated using a minibatch gradient descent on the loss:
\begin{align}
\begin{split}
\mathcal{L}_{model} &= \frac{1}{n}  \sum _{i}^{} \Vert \hat{P} \left( \phi_{s_{i}}, a_{i} \vert  \theta ^{\hat{P}} \right) - \phi_{s_{i+1}} \Vert _{2}^{2} \\
&+ \Vert \hat{R} \left( \phi_{s_{i}}, a_{i} \vert  \theta ^{\hat{R}} \right) - r_{i}^{ext} \Vert _{2}^{2}
\end{split}
\end{align}
where \textit{n} is the minibatch size,  \( \phi_{s_{i}} \)  is the state encoding,  \( r_{i}^{ext} \)  is the extrinsic reward,  \( \hat{P} \left(  \cdot,  \cdot  \vert  \theta ^{\hat{P}} \right)  \)  and  \( \hat{R} \left(  \cdot,  \cdot  \vert  \theta ^{\hat{R}} \right)  \)  are the neural networks for predicting the next state encoding and the next extrinsic reward, respectively, with parameters  \(  \theta ^{\hat{P}} \)  and  \(  \theta ^{\hat{R}} \). \par

To plan using the trained predictive model, we use model predictive control (MPC). An MPC planner observes an initial state of the world, receives an action proposal, which is a sequence of actions generated randomly or by an actor, simulates the world multiple timesteps into the future using the model, and adjusts the action proposal to optimize an objective function by backpropagation through time and gradient descent. The first action of the optimized plan is taken and then the process is repeated by replanning with the updated state information from the world in a closed loop. \par

We use a sequence of actions  \( \hat{a}_{t:t+H-1} \)  over a planning horizon \textit{H} generated by our RL actor as the action proposal and perform \textit{K} gradient descent steps on the loss:\par

\begin{equation}
\mathcal{L}_{plan} =  \left( R^{\ast}-  \sum _{h=t}^{t+H-1}\hat{r}_{h} \right) ^{2}
\end{equation}
where  \( R^{\ast} \)  is an optimal return used as a target value (usually  \( R^{\ast}=1 \) ), and  \( \hat{r}_{h}= \hat{R} \left( \hat{\phi}_{s_{h}}, \hat{a}_{h} \vert  \theta^{\hat{R}} \right)  \)  is the predicted reward at timestep \textit{h}. At each gradient descent step the plan is updated as follows:\par

\begin{equation}
\hat{a}_{t:t+H-1}^{ \left( i+1 \right) } = \hat{a}_{t:t+H-1}^{ \left( i \right) }-  \alpha _{plan}\triangledown _{\hat{a}_{t:t+H-1}^{ \left( i \right) }}\mathcal{L}_{plan}^{ \left( i \right) }
\end{equation}
where  \(  \alpha _{plan} \)  is the update rate of the model-based planner. This update is performed with much faster dynamics than the model learning update which requires physical interaction with the environment. After \textit{K} iterations, the model-based planner sends the first action of the plan  \( \hat{a}^{ \left( K-1 \right) } \)  to the agent to be executed on the environment. The optimization process of the model-based planner is shown in Fig. \ref{fig:fig2} for one iteration.\par

\subsection{Model-free learner}
\label{sec4b}
Central to our approach is the use of the expected improvement of the prediction of a trained dynamics model as an intrinsic reward to guide the exploration of the RL agent in visuomotor control tasks. Training a dynamics model directly at the pixel-level is, however, noise sensitive and involves learning task-irrelevant information. Thus, we train the dynamics model in a latent space learned with a convolutional autoencoder. Instead of using an autoencoder trained only to minimize a pixel-level reconstruction error, which gives no knowledge of what features will be useful for the desired task, we train it jointly with the \textit{Q-}function of an RL agent. This also ensures that the learned latent representation is trained on the same state distribution as the RL policy \cite{Haf181}. Fig. \ref{fig:3} shows the learning architecture of the actor and critic networks of our model-free RL agent. Any off-policy actor-critic method can be used here, such as DDPG \cite{Lil15} or CACLA \cite{Van12}. \par

\begin{figure}
    \centering
    \includegraphics[width=\linewidth]{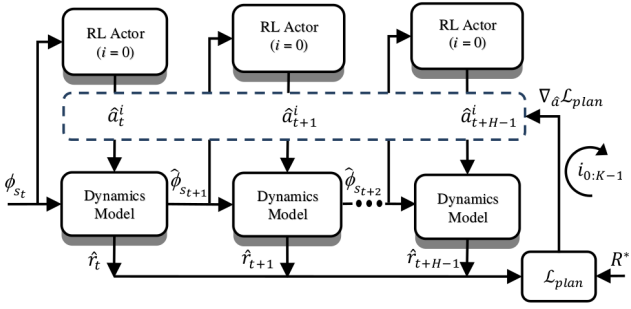}
    \caption{Model-based planner: The world is simulated \textit{H} timesteps into the future starting from an initial state of the world \(\phi_{s_t}\) and using the learned dynamics model and the action sequence generated by the RL actor. In the first iteration of the plan optimization (\textit{i}=0), the RL actor, trained simultaneously with the dynamics model, outputs an initial guess for the best action at each simulated state \(\hat{\phi}_{S_{t+1:t+H-1}}\). These actions 
    (\(\hat{a}_t^{(i=0)},\hat{a}_{t+1}^{(i=0)},\cdots,\hat{a}_{t+H-1}^{(i=0)},\)) are the first action proposal that the planner will optimize in order to minimize the loss \(\mathcal{L}_{plan}\), that is the distance between the sum of the predicted rewards \(\hat{r}_{t:t+H-1}\) and a target return \(R^*\), by backprop through time and gradient descent. This is repeated for the remaining \( K-1 \) iterations, each with an updated action proposal. At the end of the optimization process, the first action of the optimized plan \(\hat{a}^{(K-1)}\) is performed.}
    \label{fig:fig2}
\end{figure}

The convolutional autoencoder shown in Fig. \ref{fig:3}(a) is trained online to minimize the reconstruction loss at the pixel level:\par

\begin{equation}
\mathcal{L}_{rec} =  \Vert g \left( \phi_{s_{t}} \vert  \tilde{\omega}  \right) - s_{t} \Vert _{2}^{2}
\end{equation}
where  \( \phi_{s_{t}}=f \left( s_{t}  \vert   \omega  \right)  \)  is the latent encoding of the state \textit{s} at timestep \textit{t},  \( f \left(  \cdot  \vert  \omega  \right)  \)  and  \( g \left(  \cdot  \vert  \tilde{\omega}  \right)  \)  are the encoder and decoder networks with parameters  \(  \omega  \)  and  \(  \tilde{\omega}  \)  respectively. Similarly, the critic network is trained to minimize the loss:
\begin{equation}
\mathcal{L}_{critic}=  \left( y_{t} - Q \left( s_{t},a_{t} \vert  \omega, \theta ^{Q} \right)  \right) ^{2}
\end{equation}
where  \( Q \left(  \cdot, \cdot  \vert  \omega, \theta ^{Q} \right)  \)  is the critic network parameterized by  \(  \omega  \)  and  \(  \theta ^{Q} \)  and  \( y_{t}= r_{t}+  Q' \left( s_{t+1}, \mu ' \left( _{s_{t+1}} \vert  \theta ^{ \mu '} \right)  \vert  \omega ',  \theta ^{Q'} \right)  \)  is the target value with  \( Q' \left(  \cdot, \cdot  \vert  \omega ', \theta ^{Q'} \right)  \)  and  \(  \mu ' \left(  \cdot  \vert  \theta ^{ \mu '} \right)  \)  being the critic's and the actor's target networks parametrized by (\( \omega ',  \theta ^{Q'}\)) and  \(  \theta ^{ \mu '} \)  respectively. As shown in Fig. \ref{fig:3}(a), the state encoder part shared between the autoencoder and the critic is trained by minimizing the combined loss:
\begin{equation}
\mathcal{L}_{combined} =  \lambda_{rec}~\mathcal{L}_{rec} +  \lambda_{critic}~ \mathcal{L}_{critic}
\end{equation}
where $ \lambda_{rec} $ and $ \lambda_{critic} $  are weighting constants on the individual loss terms. Hence, the latent state encoding is learned to be a good state discriminator and value predictor.\par

The actor network shown in Fig. \ref{fig:3}(b) takes as input the learned latent encoding of the state and is trained according to the chosen actor-critic algorithm. This allows the actor to give a good initial action proposal to the model-based planner by using its current approximation of the optimal action at each latent state encoding generated by the latent dynamics model or received from the environment, as shown in Fig. \ref{fig:fig2}. \par

During learning, we maintain a moving window average of the prediction error of the latent dynamics model:
\begin{align}
\begin{split}
 &\langle e_{t}^{prd} \rangle = \frac{1}{ \sigma } \sum _{i=t- \sigma +1}^{t}e_{i}^{prd}\\  &\vert _{e_{i}^{prd} =  \Vert \hat{P} \left( \phi_{s_{i}}, a_{i} \vert  \theta ^{\hat{P}} \right) - \phi_{s_{i+1}} \Vert _{2}^{2}+ \Vert \hat{R} \left( \phi_{s_{i}}, a_{i} \vert  \theta ^{\hat{R}} \right) - r_{i}^{ext} \Vert _{2}^{2}}
 \end{split}
\end{align}
where  \(  \sigma  \)  is a time window and  \( e_{i}^{prd} \)  is the model prediction error at timestep \textit{i}. The average prediction error is an unbiased estimate of how unreliable the model predictions are. We also monitor the performance improvement in prediction over time by continually measuring the model learning progress:

\begin{figure}[t]
	\centering
		\includegraphics[width=3.5in,height=1.28in]{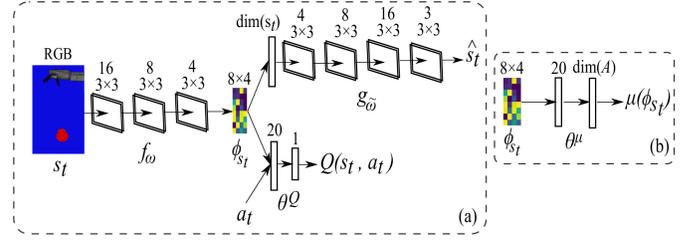}
		\caption{Model-free learner: The learning architecture consists of (a) critic and (b) actor networks. A fully convolutional autoencoder that takes in a raw image \({s_t}\) and computes a reconstruction \(\hat{s}{_t}\) is jointly trained with the critic and consists of 7 convolutional and 2 dense layers. The number and size of the convolutional filters used are shown above the corresponding layers. The actor is a feedforward network with 2 dense layers. The 32-dimensional latent representation trained by minimizing the combined critic and reconstruction loss is used as input to the actor network whose output dimensionality is dim(\textit{A}), where \textit{A} is the action space. }
		\label{fig:3}
\end{figure}


\begin{equation}\label{eq:10}
LP_{t}=  \langle e_{t-W}^{prd} \rangle  -  \langle e_{t}^{prd} \rangle 
\end{equation}
where  \( \mathcal{W} \)  is a time window. \par

The learning progress represents the change in reliability of the model and offers an informative feedback that can be used to encourage the agent to direct its exploration from states of highly predictable sensorimotor dynamics to states of less predictable dynamics. This is achieved by combining the extrinsic reward with an intrinsic reward based on the model learning progress:
\begin{equation}\label{eq:11}
r_{t}= r_{t}^{ext}+ \frac{r_{t}^{int}}{1+ D \cdot t}
\end{equation}
where  \( r_{t}^{ext} \)  is the extrinsic reward,  \( r_{t}^{int}= - LP_{t} \)  is the intrinsic reward, and  \( D>0 \)  is a decay constant used for annealing the intrinsic reward magnitude over time, since the uncertainty in the world dynamics is reduced with more directed exploration. The combined reward  \( r_{t} \)  is then used to update the critic.\par

The intrinsic reward here models the agent's curiosity to improve its knowledge of the world in situations that violate its expectation by actively seeking experiences such that the future performance improvement of its internal world model is maximized. By relying on the learning progress, the intrinsic reward is more noise-robust and suitable for non-deterministic environments where an intrinsic reward based instead on the prediction error becomes useless, as it causes the agent to focus on regions of inherently unpredictable dynamics.\par

\subsection{CMC}

At each timestep of the learning process, a standard model-free off-policy actor-critic method suggests an exploratory action that arbitrarily deviates from the actor's estimation in the hope to find and learn better actions. Similarly, a model-based planning method finds an optimal action plan by simulating the world using a predictive model with the risk of employing highly imperfect predictions. CMC presents an integrated more efficient exploration method that adaptively decides which of the model-free learner and model-based planner to query at each timestep. This decision is based on the learning progress of a latent dynamics model. When the learning progress at the previous timestep  \( LP_{t-1} \)  is positive, CMC queries the model-based planner for an optimal action (using an initial plan suggested by the model-free learner) which promises to be a better alternative than any arbitrary action. Otherwise, a negative learning progress means a high curiosity signal, which motivates the agent to select an action that improves the model. Since this curiosity is modeled by the intrinsic reward used in combination with the extrinsic reward to train the critic of the model-free learner, CMC queries the learner's actor for an optimal action. This action helps improve the learned model so that future planning with the model will become more accurate. Fig. \ref{fig:1} shows CMC with its two mutually improving components interacting with the world. The learning algorithm is summarized in Algorithm 1.


\renewcommand{\arraystretch}{1.3}
\begin{table}
\adjustbox{max width=\linewidth}{%
\begin{tabular}{rp{3.12in}}
\ChangeRT{0.9pt}
\multicolumn{2}{p{3.39in}}{{\fontsize{10pt}{12.0pt}\selectfont \textbf{Algorithm 1 }Curious Meta-Controller (CMC)}} \\
\hhline{--}
\multicolumn{1}{r}{{\fontsize{9pt}{10.8pt}\selectfont  1:}} & 
\multicolumn{1}{p{3.12in}}{{\fontsize{9pt}{10.8pt}\selectfont \textbf{Input:} Planning horizon \textit{H}, no. of plan optimization iterations \textit{\hspace*{0.5cm}K}, episode length \textit{T}, no. of episodes \textit{E}, decay constant \textit{D}. }} \\
\hhline{~~}
\multicolumn{1}{r}{{\fontsize{9pt}{10.8pt}\selectfont 2:}} & 
\multicolumn{1}{p{3.12in}}{{\fontsize{9pt}{10.8pt}\selectfont \textbf{Given:} an off-policy actor-critic method  \( \mathbb{AC} \).\  }} \\
\hhline{~~}
\multicolumn{1}{r}{{\fontsize{9pt}{10.8pt}\selectfont 3:}} & 
\multicolumn{1}{p{3.12in}}{{\fontsize{9pt}{10.8pt}\selectfont Initialize dynamics model networks  \( \hat{P} \)  and  \( \hat{R} \) }} \\
\hhline{~~}
\multicolumn{1}{r}{{\fontsize{9pt}{10.8pt}\selectfont 4:}} & 
\multicolumn{1}{p{3.12in}}{{\fontsize{9pt}{10.8pt}\selectfont Initialize actor \(   \mu  \), critic \(  Q \), target networks \(   \mu ' \)  and \(  Q' \) \  }} \\
\hhline{~~}
\multicolumn{1}{r}{{\fontsize{9pt}{10.8pt}\selectfont 5:}} & 
\multicolumn{1}{p{3.12in}}{{\fontsize{9pt}{10.8pt}\selectfont Initialize convolutional autoencoder  \(  \left( f \text{ and } g \right)  \) }} \\
\hhline{~~}
\multicolumn{1}{r}{{\fontsize{9pt}{10.8pt}\selectfont 6:}} & 
\multicolumn{1}{p{3.12in}}{{\fontsize{9pt}{10.8pt}\selectfont Initialize replay buffer \textit{R}}} \\
\hhline{~~}
\multicolumn{1}{r}{{\fontsize{9pt}{10.8pt}\selectfont 7:}} & 
\multicolumn{1}{p{3.12in}}{{\fontsize{9pt}{10.8pt}\selectfont Initialize learning progress  \( LP_{0}  \leftarrow l  \) :  \( l<0 \) }} \\
\hhline{~~}
\multicolumn{1}{r}{{\fontsize{9pt}{10.8pt}\selectfont 8:}} & 
\multicolumn{1}{p{3.12in}}{{\fontsize{9pt}{10.8pt}\selectfont \textbf{for \(  e=1~ \text{to}~ E~do \) }}} \\
\hhline{~~}
\multicolumn{1}{r}{{\fontsize{9pt}{10.8pt}\selectfont 9:}} & 
\multicolumn{1}{p{3.12in}}{{\fontsize{9pt}{10.8pt}\selectfont \ \ \  Sample initial state  \(s_{1} \) }} \\
\hhline{~~}
\multicolumn{1}{r}{{\fontsize{9pt}{10.8pt}\selectfont 10:}} & 
\multicolumn{1}{p{3.12in}}{{\fontsize{9pt}{10.8pt}\selectfont \textbf{\ \ \  for  \( t=1~\text{to}~T~do  \) }}} \\
\hhline{~~}
\multicolumn{1}{r}{{\fontsize{9pt}{10.8pt}\selectfont 11:}} & 
\multicolumn{1}{p{3.12in}}{{\fontsize{9pt}{10.8pt}\selectfont \ \ \ \ \ \ \  Compute latent state encoding  \( \phi_{s_{t}}= f \left( s_{t}  \vert   \omega  \right)  \) }} \\
\hhline{~~}
\multicolumn{1}{r}{{\fontsize{9pt}{10.8pt}\selectfont 12:}} & 
\multicolumn{1}{p{3.12in}}{{\fontsize{9pt}{10.8pt}\selectfont \textbf{\ \ \ \ \ \ \  if  \( LP_{t-1}  \geq  0 \) } \textbf{then}}} \\
\hhline{~~}
\multicolumn{1}{r}{{\fontsize{9pt}{10.8pt}\selectfont 13:}} & 
\multicolumn{1}{p{3.12in}}{{\fontsize{9pt}{10.8pt}\selectfont \ \ \ \ \ \ \ \ \ \  Query the model-based planner (see Section \ref{sec4a})}} \\
\hhline{~~}
\multicolumn{1}{r}{{\fontsize{9pt}{10.8pt}\selectfont 14:}} & 
\multicolumn{1}{p{3.12in}}{{\fontsize{9pt}{10.8pt}\selectfont \ \ \ \ \ \ \ \ \ \ \   \( a_{t}  \leftarrow  \hat{a}_{t}^{K-1}: \)   \( \hat{a}_{t}^{K-1} \)  is the optimal plan's first action}} \\
\hhline{~~}
\multicolumn{1}{r}{{\fontsize{9pt}{10.8pt}\selectfont 15:}} & 
\multicolumn{1}{p{3.12in}}{{\fontsize{9pt}{10.8pt}\selectfont \textbf{\ \ \ \ \ \ \  else}}} \\
\hhline{~~}
\multicolumn{1}{r}{{\fontsize{9pt}{10.8pt}\selectfont 16:}} & 
\multicolumn{1}{p{3.12in}}{{\fontsize{9pt}{10.8pt}\selectfont \ \ \ \ \ \ \ \ \ \ \ Query the model-free learner (see Section \ref{sec4b})}} \\
\hhline{~~}
\multicolumn{1}{r}{{\fontsize{9pt}{10.8pt}\selectfont 17:}} & 
\multicolumn{1}{p{3.12in}}{{\fontsize{9pt}{10.8pt}\selectfont \ \ \ \ \ \ \ \ \ \ \ \( a_{t}  \leftarrow   \mu  \left( \phi_{s_{t}} \vert  \theta ^{ \mu }\right), \) where \(   \mu  \)  is  \( \mathbb{AC} \)'s actor }} \\
\hhline{~~}
\multicolumn{1}{r}{{\fontsize{9pt}{10.8pt}\selectfont 18:}} & 
\multicolumn{1}{p{3.12in}}{{\fontsize{9pt}{10.8pt}\selectfont \textbf{\ \ \ \ \ \ \  end if}}} \\
\hhline{~~}
\multicolumn{1}{r}{{\fontsize{9pt}{10.8pt}\selectfont 19:}} & 
\multicolumn{1}{p{3.12in}}{{\fontsize{9pt}{10.8pt}\selectfont \ \ \ \ \ \ \  Add exploration noise  \( a_{t}  \leftarrow a_{t}+ \mathcal{N} \left( 0, 1 \right)  \) }} \\
\hhline{~~}
\multicolumn{1}{r}{{\fontsize{9pt}{10.8pt}\selectfont 20:}} & 
\multicolumn{1}{p{3.12in}}{{\fontsize{9pt}{10.8pt}\selectfont \ \ \ \ \ \ \  Execute  \( a_{t} \)  and observe  \( r_{t}^{ext} \)  and  \( s_{t+1} \) }} \\
\hhline{~~}
\multicolumn{1}{r}{{\fontsize{9pt}{10.8pt}\selectfont 21:}} & 
\multicolumn{1}{p{3.12in}}{{\fontsize{9pt}{10.8pt}\selectfont \ \ \ \ \ \ \  Compute learning progress  \( LP_{t} \), following Eq. (\ref{eq:10})}} \\
\hhline{~~}
\multicolumn{1}{r}{{\fontsize{9pt}{10.8pt}\selectfont 22:}} & 
\multicolumn{1}{p{3.12in}}{{\fontsize{9pt}{10.8pt}\selectfont \ \ \ \ \ \ \  Compute intrinsic reward  \( r_{t}^{int}= - LP_{t} \) }} \\
\hhline{~~}
\multicolumn{1}{r}{{\fontsize{9pt}{10.8pt}\selectfont 23:}} & 
\multicolumn{1}{p{3.12in}}{{\fontsize{9pt}{10.8pt}\selectfont \ \ \ \ \ \ \  Compute total reward  \( r_{t} \), following Eq. (\ref{eq:11})}} \\
\hhline{~~}
\multicolumn{1}{r}{{\fontsize{9pt}{10.8pt}\selectfont 24:}} & 
\multicolumn{1}{p{3.12in}}{{\fontsize{9pt}{10.8pt}\selectfont \ \ \ \ \ \ \  Store (\(s_{t},\phi_{s_{t}}, a_{t},r_{t},r_{t}^{ext}, s_{t+1}, \phi_{s_{t+1}}\)) in \textit{R}}} \\
\hhline{~~}
\multicolumn{1}{r}{{\fontsize{9pt}{10.8pt}\selectfont 25:}} & 
\multicolumn{1}{p{3.12in}}{{\fontsize{9pt}{10.8pt}\selectfont \ \ \ \ \ \ \  Train  \( \hat{P}, \hat{R}, f, g, \)  and  \( \mathbb{AC} \)'s  \(  \mu  \)  and \(  Q \)  on a minibatch}}  \\
\hhline{~~}
\multicolumn{1}{r}{{\fontsize{9pt}{10.8pt}\selectfont }} & 
\multicolumn{1}{p{3.12in}}{{\fontsize{9pt}{10.8pt}\selectfont \ \ \ \ \ \ \  
from \textit{R}}}\\
\hhline{~~}
\multicolumn{1}{r}{{\fontsize{9pt}{10.8pt}\selectfont 26:}} & 
\multicolumn{1}{p{3.12in}}{{\fontsize{9pt}{10.8pt}\selectfont \ \ \ \ \ \ \  Update parameters of the target networks  \(  \mu ' \)  and \(  Q'  \) }} \\
\hhline{~~}
\multicolumn{1}{r}{{\fontsize{9pt}{10.8pt}\selectfont 27:}} & 
\multicolumn{1}{p{3.12in}}{{\fontsize{9pt}{10.8pt}\selectfont \textbf{\ \ \ \ end for}}} \\
\hhline{~~}
\multicolumn{1}{r}{{\fontsize{9pt}{10.8pt}\selectfont 28:}} & 
\multicolumn{1}{p{3.12in}}{{\fontsize{9pt}{10.8pt}\selectfont \textbf{end for}}} \\
\hhline{--}
\end{tabular}}
\end{table}


\section{Experiments}
We evaluate CMC on learning continuous control tasks from raw pixels when used with different off-policy actor-critic methods. In all experiments, we use the learning architecture shown in Fig. \ref{fig:3} for approximating the policy and \textit{Q-} functions. All convolutional layers are zero-padded, have stride 1, and use ReLU activations. All dense layers use ReLU activations except for the actor's and critic's output layers that use a tanh and a linear activation respectively. The target networks' update rate  \(  \tau \)  is  \( 10^{-3} \). The loss weighting constants  $ \lambda_{rec} $  and $ \lambda_{critic} $ are set to 0.1 and 1 respectively. The dynamics model is a feedforward neural network with three dense layers: one hidden layer of 64 tanh units and two output layers for predicting the next latent encoding and reward with 32 and 1 linear units respectively. The discount factor  \(  \gamma  \)  and decay constant \( D \)  are set to 0.99 and 0.1 respectively. The time windows  \(  \sigma  \)  and  \( W \)  are set to 40 and 20 respectively. We scale the intrinsic reward to the interval  \(  \left[ -1,1 \right]  \). The planning horizon \textit{H} and the number of plan optimization iterations \textit{K }are set to 3 and 10 respectively. We train the networks using Adam optimizer \cite{Kin14} with learning rate  \( 10^{-3} \)  for the critic and the dynamics model and  \( 10^{-4} \)  for the actor and a minibatch size of 256. We perform 15 optimization steps on the critic and actor networks and 10 steps on the model network per timestep. The replay buffer size is 100K. The actor's scaled output is multiplied by a maximum step of 20 units before being sent to the model-based planner or the environment. All hyperparameters were determined empirically through preliminary experiments.\par


\begin{figure}[t]
	\centering
		\includegraphics[width=2.35in,height=1.79in]{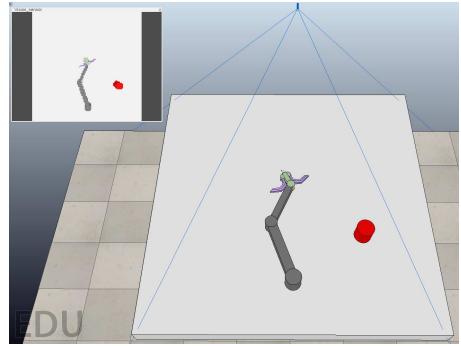}
		\caption{V-REP\ simulation environment for random target reaching. The vision sensor's output (upper-left) is fed as input to the learning algorithms.  }
		\label{fig:4}
\end{figure}


We compare the performance of DDPG and CACLA with and without CMC on learning realistic robotic reaching and grasping tasks using the V-REP robot simulator \cite{Roh13}. We run the algorithms for 10K episodes and 50 steps per episode on a single Nvidia GTX 1050 Ti 4GB GPU. \par

\subsection{Vision-based robotic reaching }
We consider random target reaching using a 3-degree of freedom (DoF) robotic arm with a two-finger gripper and a red cylinder-shaped target object. The 3D robotic environment including the vision sensor's output is shown in Fig. \ref{fig:4}.\  \par

Real-time 84$ \times $ 84 pixel RGB images from a ceiling vision sensor are used as environment states. The angular range of movement of all arm joints is  \(  \pm  \frac{ \pi }{2} \). The radius of the target zone centered around the object is one-tenth of the arm's total length and the zone area is approximately  \( 2\% \)  of the total area reachable by the arm.\par

The reward function used in the dense reward setting is:\par

 \[ r_{t}^{ext}~=   \left\{ \begin{array}{lr}
	+1 &\text{if~successful}    \\
	- \Vert c^{t}- c^{g} \Vert & \text{otherwise}  \\
	\end{array} \right. \] 
where  \(  \Vert c^{t}- c^{g} \Vert  \)  is the Euclidean distance between the centers of the target object  \( c^{t} \)  and the gripper  \( c^{g} \). In the sparse reward setting, the environment returns a reward of one when the target is reached and zero otherwise. In every episode, the position of the target object is initialized randomly within the reachable region.\par


\begin{figure}[t]
	\centering
		\includegraphics[width=3.5in,height=1.71in]{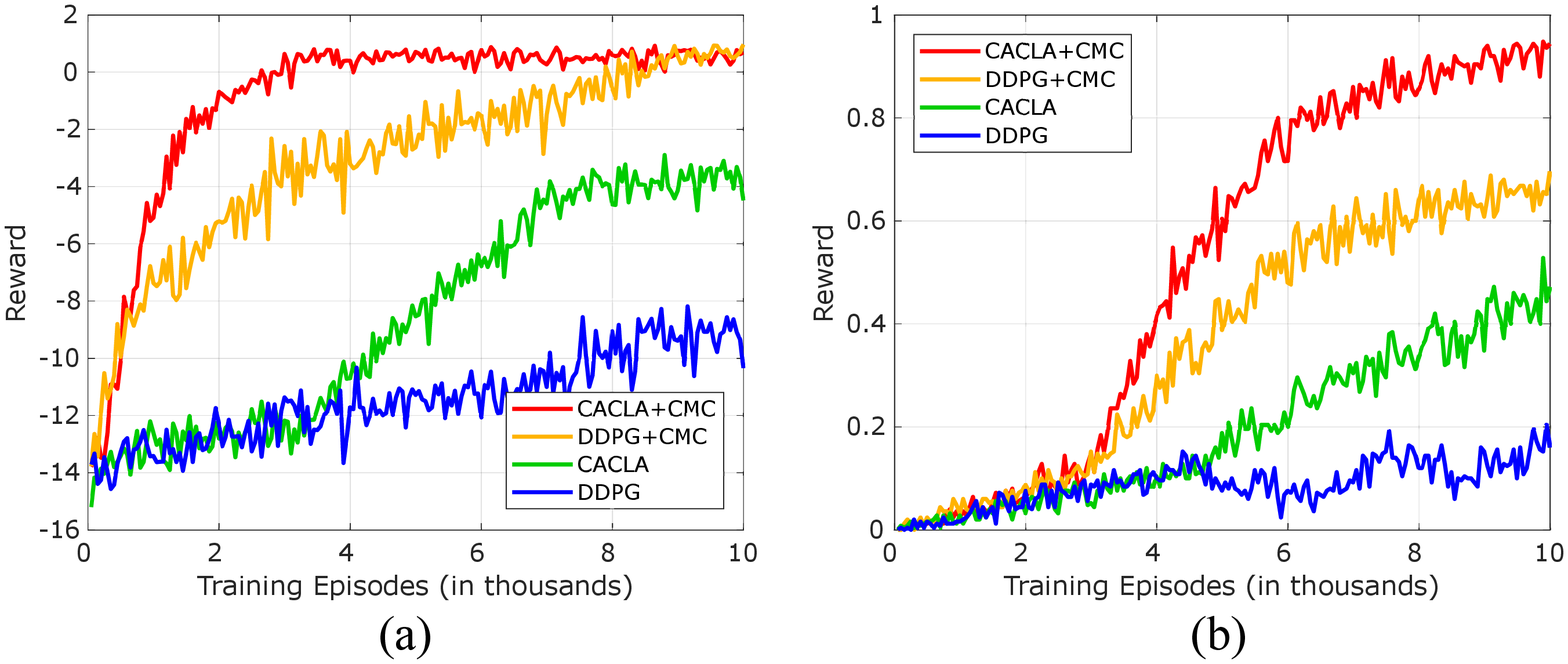}
		\caption{Learning curves of DDPG and CACLA with and without CMC on random target reaching from pixel input in two reward settings: (a) dense reward and (b) sparse reward.}
		\label{fig:5}
\end{figure}


Fig. \ref{fig:5} shows the mean episode extrinsic reward of the algorithms over 5 random seeds. DDPG and CACLA converged to policies of an episode reward of about  \( -10 \)  and  \( -4 \)  respectively in the dense reward setting (Fig. \ref{fig:5}(a)), while their CMC-based counterparts converged to near-optimal policies, with CACLA+CMC reaching a reward peak in less than 4K training episodes. In the challenging sparse reward setting, DDPG showed unstable learning with no improvement in performance and CACLA reached a poor policy of an episode reward of below 0.5, as shown in Fig. \ref{fig:5}(b). Conversely, DDPG+CMC and CACLA+CMC showed a steady increase in the episode reward, converging to 0.69 and 0.94 (i.e. \(>\) 90$\%$  success rate) respectively. \par

\subsection{Vision-based robotic grasping }
\begin{figure}[t]
	\centering
		\includegraphics[width=2.3in,height=2in]{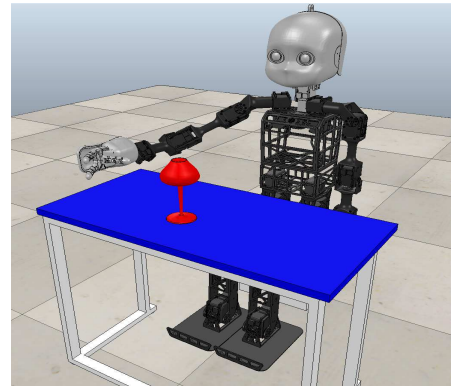}
		\caption{V-REP simulation environment for the grasping experiment, showing the NICO robot facing a table and a red glass as a grasping target.}
		\label{fig:6}
\end{figure}

In the second experiment, we evaluate the algorithms on visual robotic grasping. The need to perform multi-contact motions and to handle rigid-body collisions with a target object renders learning grasping skills more difficult than learning reaching skills. The grasping experiment here is conducted using our Neuro-Inspired COmpanion (NICO) robot \cite{Kered}. NICO is a child-sized humanoid developed at the Knowledge Technology institute, University of Hamburg, for research on cognitive neurorobotics and on human-robot interaction. Fig. \ref{fig:6} shows the V-REP simulated NICO in a sitting position in front of a table on top of which a red glass is placed and used as a target object for grasping.\par


\par

\par

\begin{figure}[!h]
	\centering
		\includegraphics[width=3.5in,height=1.51in]{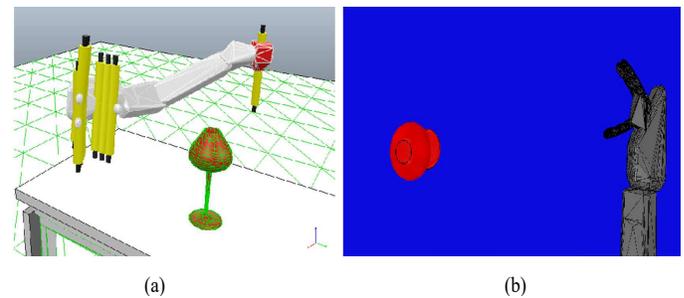}
		\caption{(a) The motor output and (b) the sensory input for the robotic grasping experiment. The axes of rotation of the controlled joints during grasp learning are depicted as yellow cylinders in (a).}
		\label{fig:7}
\end{figure}

To prevent self-collisions while also providing a large work space for learning grasping skills, we consider a control policy that involves the shoulder joint of the right arm and the finger joints of the right hand, as shown in Fig. \ref{fig:7}(a). NICO's arm has a total of 6 DoFs of which we control one in the shoulder, that has an angular range of movement of  \(  \pm  100 \)  degrees. NICO's hand is 11-DoF multi-fingered with 2 index fingers and a thumb, all of which have an angular range of movement of  \(  \pm  160 \)  degrees. The robot learns to control 2 DoFs: one for the right shoulder joint and one for the right hand (open/close). The learning algorithms take as input only the 64$ \times $ 32 pixel RGB images obtained from a vision sensor whose output is shown in Fig. \ref{fig:7}(b).\par


The reward function used in the dense reward setting is as follows:\par

 \[ r_{t}^{ext} =~  \left\{ \begin{array}{lr}
	+1 & \text{if~successful}    \\
	-1 &  \text{if object is toppled} \\
	- \Vert c^{t}- c^{h} \Vert & \text{otherwise} \\
	\end{array} \right.  \] 
where  \( c^{t} \)  and  \( c^{h} \)  are the centers of the target object and the hand respectively. To verify successful grasps, the shoulder joint is moved 20 degrees in the opposite direction to that of the last joint position with the hand closed and the distance  \(  \Vert c^{t}- c^{h} \Vert  \)  is measured. If the distance is below a threshold of 0.04 m, the last joint position update is considered successful. Otherwise, the hand is opened and the shoulder joint is moved back to its last position to complete the learning episode. In the sparse reward setting, the environment returns a zero reward for each action that does not result in the object being toppled or grasped. The target object's position is randomly changed to a new graspable position at the start of each learning episode. \textcolor[HTML]{222222}{The episode ends when the target object is grasped, toppled, or the maximum episode length \textit{T} is reached.}\par

\begin{figure}[!h]
	\centering
		\includegraphics[width=3.5in,height=1.63in]{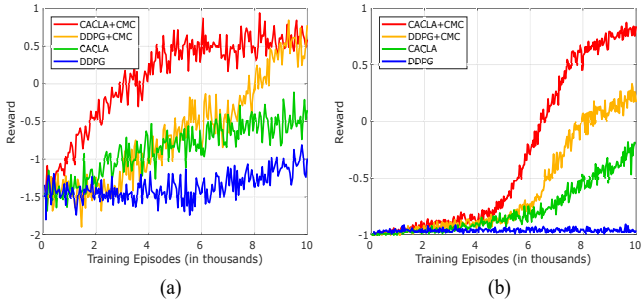}
		\caption{Learning curves of DDPG and CACLA with and without CMC on robotic grasping from pixel input in two reward settings: (a) dense reward and (b) sparse reward.}
		\label{fig:8}
\end{figure}

The mean episode extrinsic reward of running the algorithms across 5 random seeds is shown in Fig. \ref{fig:8}. All the algorithms showed no considerable performance improvement over the first 2K episodes in the dense reward setting (Fig. \ref{fig:8}(a)). Only CACLA+CMC, however, was able to converge to a policy of 0.5 episode reward in less than 5K episodes, with the other algorithms converging more slowly. The effect of CMC was more evident in the results of the sparse reward setting (Fig. \ref{fig:8}(b)). CACLA+CMC showed a sharp increase in episode reward, reaching 0.81 (81$\%$  success rate) by the end of learning, while the figure of its CACLA counterpart remained below 0. Likewise, DDPG+CMC's performance gradually improved to a policy of 0.22 episode reward, compared to its DDPG counterpart that was unable to improve its performance over the entire learning process.\par


\begin{figure}[!h]
	\centering
		\includegraphics[width=3.5in,height=1.63in]{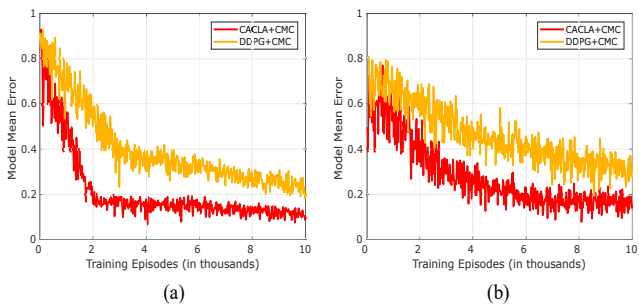}
		\caption{The performance of the latent dynamics model of CMC in the sparse reward setting: (a) on the reaching task and (b) on the grasping task.}
		\label{fig:9}
\end{figure}

Fig. \ref{fig:9} shows the average prediction error of the latent dynamics model over time, normalized to  \(  \left[ 0,1 \right]  \)  and averaged over 5 random seeds in the sparse reward setting. As shown in the figure, the error norm of the model steadily decreased in both robotic reaching and grasping tasks. This shows how the curiosity feedback drives the robot to constantly collect experiences that improve its latent dynamics model and consequently improve the model-based planner's output. The latent dynamics of the reaching environment was learned easier than 
that of the grasping environment. This is due to a higher accuracy required in the grasping task, which in turn affects the learning speed of the reward prediction part of the model.\par


\begin{figure}
    \centering
    \includegraphics[width=0.8\linewidth]{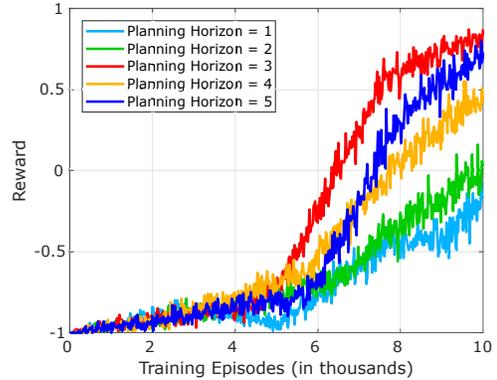}
    \caption{Learning curves of CACLA+CMC on the grasping task with sparse rewards for different planning horizons.}
    \label{fig:10}
\end{figure}

We also evaluate the effect of using different values of the planning horizon \textit{H} on the learning performance. Fig. \ref{fig:10} shows the mean episode extrinsic reward of CACLA+CMC on the grasping task with sparse rewards for different planning horizons, averaged over 5 random seeds. Going from a planning horizon of 1 to 3 steps significantly improved the learned policy. For 4-step and 5-step horizons, the performance was already close to that of the 3-step horizon, but with a slight decrease, most likely due to the last model-generated states being outside the reliable sensory region over which the learning progress is computed. \par

\section{Conclusion}
This paper introduced Curious Meta-Controller (CMC), a novel curiosity-driven controller that adaptively alternates its action choice based on either model-based planning or model-free learning. The alternation is determined by an adaptive curiosity signal based on the learning progress of a learned dynamics model. Unlike previous works, CMC considers the reliability of the model when deciding between the model-based and model-free controllers and does not require a predefined threshold to arbitrate between them. We showed that using CMC for exploration makes learning pixel-level control policies more efficient, particularly in tasks with sparse rewards. CMC can be combined with any off-policy actor-critic method, which we illustrated with DDPG and an off-policy variant of CACLA. While our approach is focused only on the action selection process, an interesting direction for future work is to investigate how to apply CMC to decide when to augment the replay buffer with model-generated experiences to learn from.
\section*{Acknowledgement}
This work was supported by the DAAD German Academic Exchange Service (Funding Programme No. 57214224) with partial support from the German Research Foundation DFG under project CML (TRR 169).
\bibliographystyle{IEEEtran}
\bibliography{main}

\end{document}